\documentclass[10pt,twocolumn,letterpaper]{article}

\usepackage{cvpr}              

\usepackage{graphicx}
\usepackage{amsmath}
\usepackage{amssymb}
\usepackage{booktabs}

\usepackage[pagebackref,breaklinks,colorlinks]{hyperref}

\usepackage[accsupp]{axessibility}  

\usepackage[capitalize]{cleveref}
\crefname{section}{Sec.}{Secs.}
\Crefname{section}{Section}{Sections}
\Crefname{table}{Table}{Tables}
\crefname{table}{Tab.}{Tabs.}

\def\cvprPaperID{45} 
\def\confName{CVPR}
\def\confYear{2023}

\begin{document}

\title{AUD-TGN: Advancing Action Unit Detection with Temporal Convolution and GPT-2 in Wild Audiovisual Contexts }


\author{
Jun Yu$^1$, Zerui Zhang$^1$, Zhihong Wei $^1$\thanks{Corresponding author}, Gongpeng Zhao$^1$, Zhongpeng Cai$^1$, Yongqi Wang$^1$,\\ Guochen Xie$^1$, Jichao Zhu$^1$, Wangyuan Zhu$^1$  \\
$^1$University of Science and Technology of China\\
\tt\small harryjun@ustc.edu.cn\\
\tt\small \{igodrr,weizh588,zgp0531,zpcai,wangyongqi,xiegc,\\
\tt\small jichaozhu,zhuwangyuan\}@mail.ustc.edu.cn 
}

\maketitle

\begin{abstract}
\normalfont
Leveraging the synergy of both audio data and visual data is essential for understanding human emotions and behaviors, especially in in-the-wild setting. 
Traditional methods for integrating such multimodal information often stumble, leading to less-than-ideal outcomes in the task of  facial action unit detection. 
To overcome these shortcomings, we propose a novel approach utilizing audio-visual multimodal data. 
This method enhances audio feature extraction by leveraging Mel Frequency Cepstral Coefficients (MFCC) and Log-Mel spectrogram features alongside a pre-trained VGGish network. 
Moreover, this paper adaptively captures fusion features across modalities by modeling the temporal relationships, and ultilizes a pre-trained GPT-2 model for sophisticated context-aware fusion of multimodal information. 
Our method notably improves the  accuracy of AU detection by understanding the temporal and contextual nuances of the data, showcasing significant advancements in the comprehension of intricate scenarios.
These findings underscore the potential of integrating temporal dynamics and contextual interpretation, paving the way for future research endeavors.

\end{abstract}

\section{Introduction}
\label{sec:intro}


The sixth Competition on Affective Behavior Analysis in-the-wild (ABAW6) \cite{kollias20246th,kollias2023abaw2,kollias2023multi,kollias2023abaw,kollias2022abaw,kollias2021analysing,kollias2021affect,kollias2021distribution,kollias2020analysing,kollias2019expression,kollias2019deep,kollias2019face,zafeiriou2017aff } targets challenges in analyzing human emotions through facial expressions. Facial Action Units (AUs), fundamental for expressing emotions, are the focus of significant research due to their communicative importance \cite{kollias2022abaw, kollias2023abaw}. Derived from the Facial Action Coding System (FACS) \cite{prince2015facial}, AUs are critical for a variety of applications, from psychology to security. However, detecting AUs accurately, especially in uncontrolled environments, is complex due to diverse expressions and the necessity for multimodal data integration. Our work responds to this challenge, aiming to refine AU detection methods and explore novel multimodal fusion techniques for a deeper understanding of emotional expressions.

The analysis of facial action units (AUs), essential for interpreting human emotions and expressions, relies on the Facial Action Coding System (FACS) to associate specific AUs with localized facial regions. Traditional methods for detecting AUs utilized handcrafted features to represent these regions\cite{chu2013selective,ding2013facial,eleftheriadis2015multi,liu2013aware,wang2013capturing}, laying the groundwork for this field of study. However, these methods were limited in their adaptability to the wide range of facial expressions, often unable to accurately capture the nuances of facial movements or adjust to facial posture changes. The introduction of deep learning marked a new era for AU detection, with deep neural networks offering more sophisticated means for extracting and analyzing facial features. Techniques such as employing face landmarks or segmenting aligned faces into patches have been developed to more accurately locate facial areas related to AUs, yet these approaches often fixedly extracted facial features, limiting their effectiveness. To address these limitations, recent innovations have introduced more flexible and adaptive strategies. For instance, one method employed a three-stage training strategy to enable encoders to adaptively extract features related to facial local regions, although this necessitated additional annotations for face landmarks and depended on multi-task learning\cite{tang2021piap}. Furthermore, recognizing that the activation of AUs is not isolated but interconnected, recent studies have utilized graph neural networks to explore the relationships between AUs, employing a two-stage training strategy to capture multi-dimensional edge features that reflect the complex web of AU interactions\cite{luo2022learning}. However, the complexity of such training strategies has highlighted the need for a more streamlined approach to AU detection.

This study initiates with the preprocessing of video data to dissect audio and visual streams, whereupon Log-Mel\cite{meng2019speech} spectrogram and Mel Frequency Cepstral Coefficients (MFCC)\cite{tiwari2010mfcc} are extracted for audio. Subsequent to this foundational step, we leverage pre-trained VGG\cite{simonyan2014very} and ResNet architectures\cite{he2016identity} to distill intricate audio and visual features on a per-frame basis. To circumvent the challenges posed by the temporal continuity of video and the homogeneity among audio frames, our methodology incorporates dilated convolutional layers\cite{li2018csrnet}, thereby augmenting the model's capacity for temporal context capture and enriching the extraction of temporal features across distinct input branches. Following feature extraction, concatenation and convolutional operations facilitate the integration of these multimodal inputs. Crucially, the incorporation of a pre-trained GPT-2\cite{lagler2013gpt2} model, with its sophisticated context-aware attention mechanism, marks a pivotal phase in our approach, enhancing the discernment of nuanced facial expressions and their evolution throughout the video sequence\cite{zhou2024one}. This rigorously structured framework, which transitions from initial data preprocessing to the application of advanced neural networks, provides a robust strategy for interpreting complex emotional and behavioral cues within video data, underscoring the transformative potential of deep learning in the domain of affective computing.


\begin{figure*}
    \includegraphics[width=\textwidth]{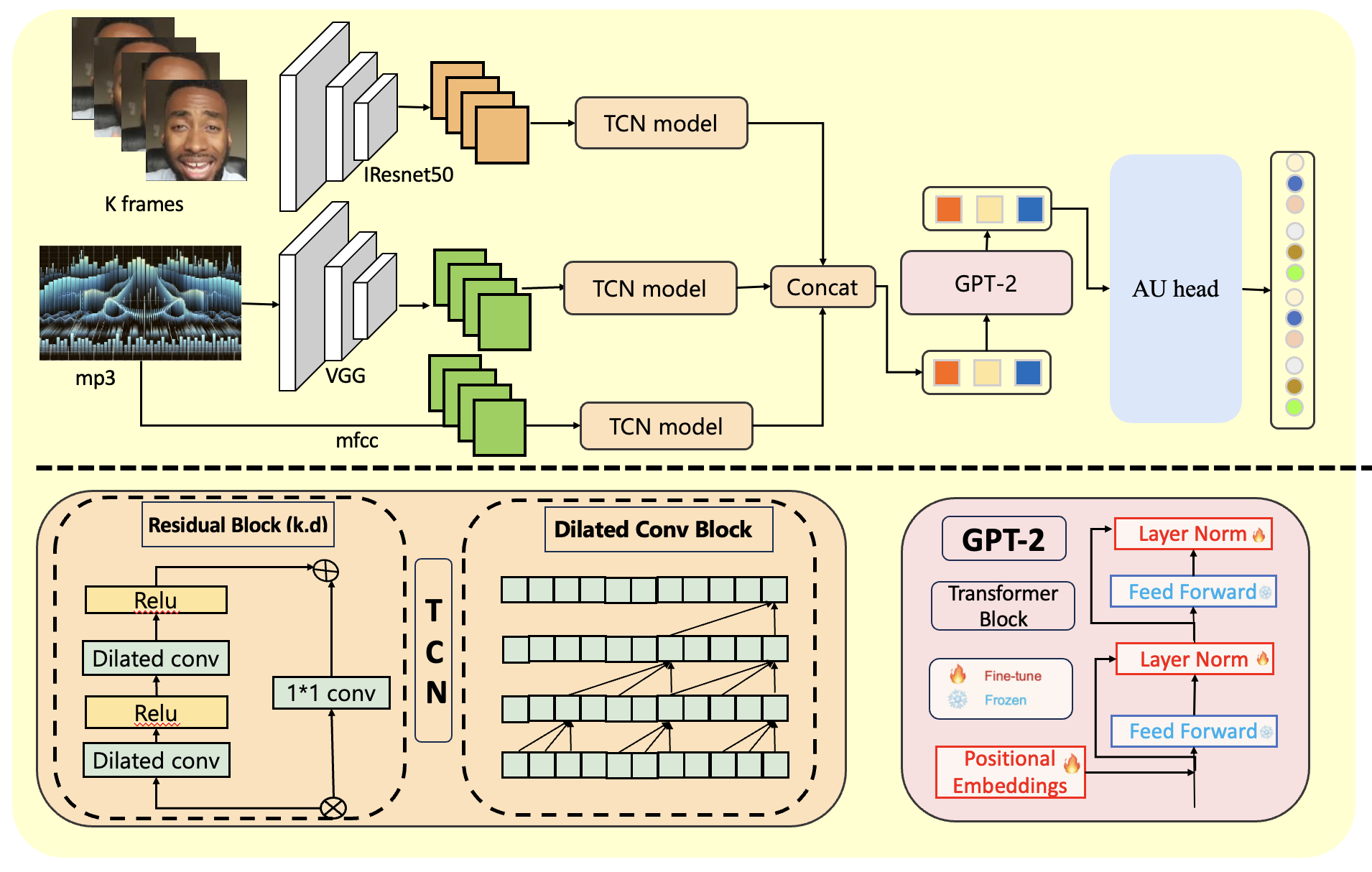}
    \caption{The flowchart presents a multimodal approach for detecting facial action units, employing pre-trained iResnet50 networks for initial feature extraction from video and audio, which are then refined through Temporal Convolutional Networks to capture the temporal dynamics. These features are integrated via a fine-tuned GPT-2 model before being classified by an AU detection head. The detailed submodules illustrate the internal workings of the TCN, emphasizing its dilated convolution blocks for expansive temporal feature capture, and the GPT-2 model, highlighting the transformer mechanism and fine-tuning approach that enables contextual understanding of the features.}
    \label{fig:pipeline}
\end{figure*}

To sum up, our contributions  can be summarized as:
\begin{itemize}
    \item We streamline AU detection by preprocessing video into audio and visual streams, extracting Log-Mel and MFCC features, and utilizing pre-trained VGG and ResNet for advanced feature extraction.
    \item Our method incorporates dilated convolutional layers to enhance temporal context capture, addressing video's temporal continuity and audio frame homogeneity.
    \item We employ a pre-trained GPT-2 model for its context-aware attention mechanism, significantly improving the detection and interpretation of nuanced facial expressions throughout video sequences.
\end{itemize}

\section{Related Work}
\label{sec:Relatedwork}
Addressing the complexities of facial action unit (AU) detection, the field confronts notable challenges including the limited identity variance in prevalent datasets and the extraction of pertinent local features for each AU. Traditional methods have shown substantial limitations\cite{benitez2017recognition, jacob2021facial, kollias2019face}, particularly those dependent on manual feature specification, due to the intricate and nuanced nature of AU annotations.

To overcome these barriers, recent efforts have embraced additional facial landmarks to delineate important local features and have utilized multi-task learning to refine AU detection model efficacy. For example, the SEV-Net \cite{yang2021exploiting} model generates local region attention maps through textual descriptors, providing a novel lens to focus on salient facial areas. Similarly, Tang et al. \cite{tang2021piap} implement a three-stage training strategy, drawing upon facial landmark information in a multi-task learning paradigm to direct the model's attention to key facial regions.

Nevertheless, such approaches often necessitate the inclusion of supplementary landmark annotations, and they may neglect the complex associations between AUs. Luo et al. \cite{luo2022learning} propose a method using a graph neural network, employing a two-stage training strategy to capture the relational state between AUs. Yet, this method defaults to using simple fully connected layers to represent each AU node, foregoing the additional landmark annotations and requiring an initial training phase for the network to effectively learn node information. These techniques, despite their advances, highlight an ongoing quest in the field to develop models that can autonomously learn critical facial features and their interdependencies without heavy reliance on external annotations.

\section{Method}
In this section, we will describe our proposed approach in detail. As shown in Figure \ref{fig:pipeline}, Our methodology for facial action unit (AU) detection commences with preprocessing video into audio and visual streams. For visual features, images $I_v \in \mathbb{R}^{H \times W \times 3}$ are input into a ResNet model\cite{duta2021improved} pre-trained on Glint360K\cite{an2021partial} , producing features $F_v \in \mathbb{R}^{H' \times W' \times C_v}$, where only the final layer is updated. Audio features are extracted from Log-Mel spectrograms $I_a \in \mathbb{R}^{T \times F}$ through a pre-trained VGGish network, combined with MFCC, resulting in $F_a \in \mathbb{R}^{1 \times C_a}$. Temporal dynamics are captured via TCN\cite{bai2018empirical}, yielding $T_v \in \mathbb{R}^{1 \times C_t}$ for visual and $T_a \in \mathbb{R}^{1 \times C_t}$ for audio features. Fusion of these. temporal features through a Transformer network generates a comprehensive representation $F_{fusion} \in \mathbb{R}^{1 \times C_f}$, subsequently processed by a multi-class classifier to predict AU presence $P_{AU}$. This streamlined approach leverages deep learning to efficiently detect AUs, integrating complex audio-visual data.

\subsection{Data preprocess}
In our data preprocessing pipeline, we meticulously prepare both visual and auditory inputs to ensure that they are optimally primed for feature extraction. For the visual component, each frame of the video sequence is processed through a ResNet network that has been pre-trained on a facial dataset. This pre-training allows the network to generate high-fidelity representations that are particularly attuned to facial features, which are crucial for AU detection. Let $I_{frame}$ denote the input frame and $F_{ResNet}$ represent the output feature vector obtained from ResNet:
\begin{equation}
F_{ResNet} = ResNet(I_{frame}) \in \mathbb{R}^{H' \times W' \times C_{v}}
\end{equation}
Here, $H'$ and $W'$ denote the height and width of the processed feature maps, while $C_{v}$ denotes the number of channels.

Moving on to the auditory aspect, we begin by extracting two types of features: the Mel Frequency Cepstral Coefficients (MFCC) and Log-Mel spectrogram features. These features are particularly effective in capturing the essence of sound and are fundamental to a variety of audio processing tasks. The Log-Mel features are then passed through a VGGish network, which has been pre-trained to encode these features into a robust auditory representation known as VGGish features:
\begin{equation}
F_{VGGish} = VGGish(LogMel(I_{audio})) \in \mathbb{R}^{C_{a}}
\end{equation}
In this formula, $LogMel(I_{audio})$ refers to the Log-Mel spectrogram features of the audio input $I_{audio}$, and $F_{VGGish}$ denotes the encoded VGGish features with $C_{a}$ being the feature dimensionality. This pre-trained VGGish model serves to effectively distill the audio information into a format that is conducive to our subsequent multimodal analysis, enabling a more comprehensive understanding of the auditory signals associated with the video data.

This dual-faceted preprocessing approach sets a robust foundation for the ensuing stages of our facial action unit detection framework, ensuring that both the visual and auditory modalities are represented with high granularity and are well-suited for the deep learning tasks ahead.

\subsection{TCN}
In our methodological framework for temporal feature processing, video sequences are segmented into clips each comprising 200 consecutive frames to prepare for Temporal Convolutional Network (TCN) application. The TCN leverages dilated convolutions to process temporal sequences efficiently, enhancing the model's ability to capture broader contextual information without a commensurate increase in computational demand.

Dilated convolutions enable the network to have an exponentially larger receptive field, which is crucial for incorporating long-range temporal dependencies. For an input sequence $X \in \mathbb{R}^{L \times C_{in}}$, where $L=200$ is the sequence length and $C_{in}$ is the number of input channels, the TCN applies a dilated convolution operation to produce an output sequence $Y \in \mathbb{R}^{L \times C_{out}}$, with $C_{out}$ as the number of output channels. The dilation factor $d$ determines the spacing between the kernel's elements. The dilated convolution operation in TCN, parameterized by weights $\theta$, can be expressed as:

\begin{equation}
Y(t) = (X *_d f)(t) = \sum_{s=0}^{k-1} f(s) \cdot X(t - d \cdot s)
\end{equation}

where $*_d$ denotes the dilated convolution operation, $f$ represents the filter of size $k$, and $t$ indexes the time step. The dilation factor $d$ allows the filter to cover a wider span of the input sequence per time step, effectively enlarging the receptive field and enabling the capture of temporal patterns significant for AU detection.



\subsection{Leveraging Pretrained Transformer GPT-2}
Upon fusing the temporal features derived from each modality via TCN networks, we consolidate these into a unified feature vector $F_{concat} \in \mathbb{R}^{BL \times 3C}$. A convolutional layer further refines this vector, ensuring enhanced multimodal data integration and yielding $F_{conv} \in \mathbb{R}^{BL \times C'}$:

\begin{equation}
F_{concat} = Concat(T_v, T_a, \ldots)
\end{equation}
\begin{equation}
F_{conv} = Conv(F_{concat}) \in \mathbb{R}^{BL \times C'}
\end{equation}

Central to our framework is the pre-trained GPT-2 model, which, despite being largely frozen to preserve its linguistic feature extraction prowess, has been fine-tuned specifically for our task—most notably, the Layer Normalization (LN) components. In the final stages of our pipeline, we selectively utilize only four modules of GPT-2, which effectively contextualize $F_{conv}$ and enrich it with a detailed depiction of the inter-token relationships and temporal dependencies:

\begin{equation}
F_{GPT2} = GPT2_{\text{4-modules, LN-finetuned}}(F_{conv}) \in \mathbb{R}^{BL \times N}
\end{equation}

The resultant feature representation $F_{GPT2}$, enhanced by the GPT-2 modules, encapsulates critical contextual information along the temporal dimension. When subjected to a classification head, this information facilitates the accurate prediction of each AU's presence. The strategic employment of a subset of GPT-2's transformer modules demonstrates the significant benefits of leveraging targeted components of pre-trained NLP models to capture the intricate subtleties of facial expressions within affective computing.





\section{Loss Function and Metric}
To train our model for the task of facial action unit (AU) detection, we employ the Binary Cross-Entropy (BCE) loss, which for a single instance is defined as:
\begin{equation}
    L_{AU\_CE} = - \frac{1}{N} \sum_{i=1}^{N} W_{au_i} \left[y_i \log(\hat{y}_i) + (1 - y_i) \log(1 - \hat{y}_i)\right]
\end{equation}
where $N$ is the number of classes (action units), $y_i$ is the binary label for the $i$-th AU (1 for presence and 0 for absence), $\hat{y}_i$ is the predicted probability of the $i$-th AU being present, and $W_{au_i}$ is the weight associated with the $i$-th AU to address class imbalance.

For evaluating our model's performance, we compute the F1 score for each AU class, and then calculate the mean F1 score across all AUs as follows:
\begin{equation}
    \text{F1}_{\text{mean}} = \frac{1}{N} \sum_{i=1}^{N} \text{F1}_i
\end{equation}
The F1 score for each class is the harmonic mean of precision and recall, providing a balance between the two for a robust assessment of model performance.





\subsection{Post-Process}
Since Aff-Wild2 \cite{zafeiriou2017aff,kollias2019deep, kollias2019face, kollias2019expression, kollias2020analysing,kollias2021affect,kollias2021analysing,kollias2021distribution,kollias2022abaw, kollias2023abaw,kollias2023abaw2}  dataset is After obtaining the prediction confidence levels for each category, we employ a thresholding technique to segregate the results. The inherent label imbalance in our dataset, predominantly skewed towards the '0' class, results in generally lower confidence scores, not uniformly distributed between 0 and 1. To counteract this and enhance our model's performance, we adjust the threshold for classification. A lower threshold value was empirically found to significantly improve the scoring metrics. Consequently, we systematically explored a range of threshold values on the validation set to identify the optimal solution, thereby optimizing our model's performance in the face of label imbalance.

\begin{table*}
  \centering
  \caption{Ablation study results on the official validation set, the highest score is indicated in bold.}
  \label{tab:table1}
  \resizebox{\linewidth}{!}{
  \begin{tabular}{ccccccccccccccc}
    \toprule
    \textbf{Method} & \textbf{pretrained-resnet} & \textbf{resnet-finetune} & \textbf{tcn} & \textbf{GPT-2} & \textbf{Post-Process} & \textbf{F1 Score (\%)} \\ 
    \midrule
	baseline & & &  & & &36.5 \\

	pretrained & $\checkmark$ & &  & & &42.8 \\
 
    pretrained+finetune & $\checkmark$ & $\checkmark$ & & &  &44.5 \\
 
    TCN & $\checkmark$ & $\checkmark$ & $\checkmark$& & &42.6 &  \\

    GPT-2 & $\checkmark$ & $\checkmark$  &  & $\checkmark$& &48.9 \\
	
    TCN+GPT-2 & $\checkmark$ & $\checkmark$ & $\checkmark$ & $\checkmark$&  &51.4 \\

    TCN+GPT-2 & $\checkmark$ & $\checkmark$ & $\checkmark$ & $\checkmark$ &  $\checkmark$& \textbf{53.7} \\
    
    \bottomrule
  \end{tabular}
  }
\end{table*}

	
    

\begin{table}[ht]
  \centering
  \caption{The average F1 scores (in \%) of different teams on the official Aff-wild2 validation set and test set. Our results are indicated in bold. The last line represent our best results in the post challenge evaluation phase.}
  \label{tab:table2}
  \resizebox{\linewidth}{!}{
    \begin{tabular}{lc}
      \toprule
      \textbf{Teams} & \textbf{F1 Score (\%)} \\
      \midrule
      Netease Fuxi Virtual Human \cite{zhang2023facial} & \textbf{55.27} \\
      SituTech & 54.04 \\
      USTC-IAT-United \cite{yu2023local} & 50.98 \\
      PRL & 49.14 \\
      SZFaceU \cite{wang2023spatiotemporal} & 47.93 \\
      HSE-NN-SberAI \cite{savchenko2023emotieffnet} & 48.21 \\
      CtyunAI \cite{Zhou_2023} & 48.18 \\
      HFUT-MAC \cite{zhang2023facial} & 46.88 \\
      SCLAB CNU \cite{nguyen2023transformerbased} & 45.63 \\
      USTC-AC \cite{wang2023facial} & 43.29 \\
      USC IHP \cite{yin2023multimodal} & 42.91 \\
      ACCC & 37.76 \\
      baseline \cite{kollias2023abaw} & 36.5 \\
      USTC-IAT-United (Ours Best) & \textbf{53.7} \\
      \bottomrule
    \end{tabular}
  }
\end{table}

\section{Experiment}
In this section, we will provide a detailed description of the used datasets, the experiment setup, and the experimental results.
\subsection{Datasets}
\label{sec:dataset}
\textbf{AU Datasets.}
The Aff-Wild2 dataset, a substantial extension of the original Aff-wild1 repository, stands at the forefront of affective behavior analysis, offering an unprecedented breadth of annotated data. Spanning 567 videos annotated for valence-arousal dynamics, and 548 videos each for eight distinct expression categories, this dataset encompasses a comprehensive range of human emotions. Additionally, 547 videos meticulously annotated for 12 distinct Action Units (AUs) enhance the dataset's granularity. The dataset is further bolstered by a collection of 172,360 images annotated across the valence-arousal spectrum, six basic expressions plus neutral and 'other' states, and 12 AUs, providing a multifaceted view of human affect.

The Action Unit Detection task, a critical component of the dataset, is represented in 548 videos that capture the six fundamental expressions, the neutral state, and an 'other' category encapsulating affective states beyond the basic emotions. With close to 2.6 million frames and contributions from 431 diverse participants (265 males and 166 females), the dataset's depth is unparalleled, annotated with precision by a team of seven experts. Aff-Wild2 stands as a testament to spontaneous human affect in naturalistic settings, propelling affective computing closer to the complexities of real-world scenarios.

\subsection{Training details}
For the training phase of our study, we employed a pre-trained iResNet network. The fine-tuning was confined to the network's final layer parameters, adjusted to a learning rate that is one-tenth of the standard rate. The optimization process was guided by the AdamW optimizer, spanning a duration of 50 epochs.In our training strategy, a crucial element is the optimization of the learning rate schedule. We adopt a linear warmup strategy that begins with an initial rate and linearly increases to reach a learning rate of 0.0001 within the span of 2000 iterations. This gradual increment allows the model to adjust to the complexity of the task, ensuring stable convergence.If there is no improvement on the validation set for five consecutive epochs, the learning rate is scaled down to a tenth of its value. This approach aims to fine-tune the learning process adaptively based on the model's performance.
Our model processed video segments with a length of 200 frames each, and we set the batch size to 4. Notably, during the training phase, the 30th epoch marked a milestone as we observed the best performance on the validation set at this point.
This methodical training regimen, marked by strategic learning rate adjustments and careful monitoring of validation performance, underscores our commitment to achieving a robust model that reliably understands and classifies affective behaviors as manifested in the Aff-Wild2 dataset.


\subsection{Results}
\textbf{Validation Set Results.} The average F1 scores (in \%) of different teams on the official Aff-wild2 validation set are shown in Table. \ref{tab:table2}. Our method achieves good performance (53.7\%) on the official validation set. In fact, this score is also quite high among all the participating teams, indicating to some extent the good potential of our approach. In addition, more discussion of the validation set results can be found in Sec. \ref{sec:ablation study}.


\subsection{Ablation Study}
\label{sec:ablation study}
A rigorous ablation study was conducted to ascertain the contributory significance of various model components to our AU detection task. We examined the impact of a pre-trained iResNet, the fine-tuning of the network's last layer, the implementation of Temporal Convolutional Networks (TCN), and the inclusion of a pre-trained GPT-2 model.

The removal of the pre-trained iResNet precipitated a marked decrease in performance, underscoring its integral role in discerning complex patterns vital for AU detection. The model's accuracy rose from an initial 36.5\% to 42.8\% with the integration of iResNet.

Implementing TCN for temporal feature extraction further enhanced our model's performance, boosting accuracy from 44.5\% to 46.6\%. This underscores the significance of temporal dynamics in AU detection.

Subsequently, fine-tuning the network's last layer contributed positively, raising performance from 42.8\% to 44.5\%. This fine-tuning process proves critical in adapting the network to the nuances of our specific dataset.

Moreover, the inclusion of a pre-trained GPT-2 model resulted in a notable increase in accuracy, from 44.5\% to 48.9\%, showcasing the effectiveness of advanced NLP models in encoding the temporal and contextual subtleties of facial expressions.

Collectively, the ablation study results highlight the essential roles of pre-trained networks, temporal feature extraction with TCN, and strategic fine-tuning in improving AU detection, particularly in complex, real-world conditions.


\section{Conclusion}
In our research, we've demonstrated the effectiveness of integrating Temporal Convolutional Networks (TCN) with pre-trained iResNet and GPT-2 models for the nuanced task of facial action unit (AU) detection in "in-the-wild" settings. By leveraging TCN for dynamic feature extraction and enriching feature representation through pre-trained models, we achieved notable improvements in AU detection accuracy. The results underscore the synergistic impact of combining temporal analysis with advanced neural architectures in enhancing affective computing applications.

{\small
\bibliographystyle{ieee_fullname}
\bibliography{egbib}
}

\end{document}